%% file: main.tex
\documentclass{article} % For LaTeX2e
\usepackage{iclr_set,times}

\input{math_commands.tex}

\usepackage{hyperref}
\usepackage{url}
\usepackage{graphicx}
\usepackage{multirow}
\usepackage{booktabs}
\definecolor{forest}{RGB}{44, 154, 32}

\title{Sequence-Level Certainty Reduces Hallucination In Knowledge-Grounded Dialogue Generation}

\author{Yixin Wan\\
University of California, Los Angeles\\
\texttt{elaine1wan@cs.ucla.edu}
\And
Fanyou Wu \& Weijie Xu \& Srinivasan H. Sengamedu \\
Amazon Science\\
\texttt{\{fanyouwu, weijiexu, sengamed\}@amazon.com}
}

\newcommand{\tableblank}[1]{\hspace*{#1}\linebreak[0]}
\iclrfinalcopy % Uncomment for camera-ready version, but NOT for submission.
\begin{document}

\maketitle

\vspace{-0.8em}
\begin{abstract}
\vspace{-0.6em}
In this work, we propose sequence-level certainty as a common theme over hallucination in Knowledge Grounded Dialogue Generation (KGDG).
We explore the correlation between the level of hallucination in model responses and two types of sequence-level certainty: probabilistic certainty and semantic certainty.
Empirical results reveal that higher levels of both types of certainty in model responses are correlated with lower levels of hallucination. 
We further propose Certainty-based Response Ranking (CRR), a decoding-time hallucination mitigation method that samples several response candidates, ranks them based on sequence-level certainty, and outputs the response with the highest certainty level.
Aligning with our definitions of sequence-level certainty, we design $2$ types of CRR approaches: Probabilistic CRR (P-CRR) and Semantic CRR (S-CRR).
P-CRR ranks individually sampled model responses using the arithmetic mean log-probability of the entire sequence.
S-CRR approaches certainty estimation from meaning-space, and ranks model response candidates based on their semantic certainty level as measured by an entailment-based Agreement Score (AS).
Through extensive experiments across $3$ KGDG datasets, $3$ decoding methods, and $4$ KGDG models, we validate the effectiveness of CRR for reducing hallucination in KGDG task.
\end{abstract}

\vspace{-1.5em}
\section{Introduction}
\vspace{-0.5em}
Previous works have researched the problem of hallucination in Knowledge-Grounded Dialogue Generation (KGDG) task~\citep{li-etal-2019-incremental, shuster-etal-2021-retrieval-augmentation, Santhanam2021RomeWB, honovich-etal-2021-q2, dziri-etal-2022-evaluating, rashkin-etal-2021-increasing}.
For KGDG, a dialogue model is given a piece of textual knowledge and a series of conversation history, and is expected to generate informative and meaningful responses to the previous dialogue with the provided knowledge~\citep{li-etal-2022-knowledge}.
A model response is therefore defined to be ``hallucinated'' if it is inconsistent or unsupported by the knowledge given in the model input~\citep{filippova-2020-controlled, dziri2022faithdial}.

% Dialogue models still struggle to generate faithful responses on the KGDG task~\citep{honovich-etal-2021-q2}.
% Understanding and mitigating hallucination in KGDG models therefore remains an important research question in the domain of NLG.
% Previous work's~\cite{xiao-wang-2021-hallucination} exploration of the correlation between uncertainty and hallucination is limited to the study of token-level uncertainty on Image Captioning (IC) task.
% Among previous works, \citeauthor{xiao-wang-2021-hallucination} investigated model hallucination on Image Captioning (IC) task.
% They proposed to use token-level probabilistic entropy to measure predictive uncertainty, and were the first to show that higher predictive uncertainty is associated with higher chance of hallucination.
% However, the proposed probability-based estimation of model uncertainty in their study remains on token-level, and therefore neglects sequence-level semantic information that might also indicate uncertainty in model generations.
% Token-level probability-based predictive uncertainty is therefore not a satisfactory and generalizable common theme across model hallucinations in NLG.
Our work proposes and investigates \textbf{sequence-level certainty} as a general common theme over hallucinations in KGDG.
We dissect sequence-level model certainty into two categories: \textbf{probabilistic certainty} and \textbf{semantic certainty}.
% In addition, we demonstrate through statistical proof that semantic certainty is a good estimator to probabilistic certainty.
% This proves semantic certainty as a promising alternative to probabilistic certainty in black-box scenarios, where model output probabilities are not always available.
To measure semantic certainty, our study proposes \textbf{Agreement Score (AS)}, which is defined as the overall level of semantic entailment of each candidate with all other candidates.
We first prove through experiments that higher levels of both types of certainty are correlated with lower levels of hallucination in model outputs.
Furthermore, we propose Certainty-based Response Ranking (CRR) to mitigate the hallucination of KGDG models during decoding time.
Specifically, aligning with our categorization of sequence-level certainty, we establish $2$ types of CRR approaches: \textbf{Probabilistic CRR (P-CRR)}, and \textbf{Semantic CRR (S-CRR)}.
P-CRR simply ranks several independently sampled model responses by their probabilistic certainty, measured by the arithmetic mean log-probability over entire sequences.
% , which we follow previous work to establish as the arithmetic mean log-probability over entire sequences \cite{kuhn2023semantic,murray2018correcting}.
S-CRR approaches certainty estimation from a semantic perspective, and ranks various independently sampled model response candidates by their semantic certainty.

% We first empirically show that both a higher level of probabilistic certainty and a higher level of semantic certainty are positively and significantly correlated with a lower level of hallucination through comprehensive experiments across $4$ models on KGDG task.
% Then, we 
We validate the effectiveness of our P-CRR and S-CRR methods through extensive experiments on $3$ KGDG datasets, $3$ different decoding methods, and $4$ KGDG models with varied sizes.
% , with $3$ different decoding methods.
Experiment results demonstrate that both P-CRR and S-CRR significantly reduce hallucinations in model outputs across all experiment settings.
% We conduct ablation experiments to further investigate how the number of response candidates sampled for ranking can influence performance of CRR methods.
% Interestingly, we observe that S-CRR achieves significant improvement with increased number of response candidates, whereas P-CRR maintains similar performance.
% This indicates that S-CRR better captures semantic certainty in model generations through more pair-wise alignment analysis between additional response candidates, which further proves the effectiveness of the S-CRR method.
Our work provides novel and significant findings on the relationship between sequence-level certainty and hallucination on KGDG task, opening up a new direction for future research to further explore and understand the hallucination phenomenon.
 \vspace{-0.3cm}  
 
\begin{figure}[t]
\vspace{-1.2em}
\includegraphics[width=0.8\linewidth]{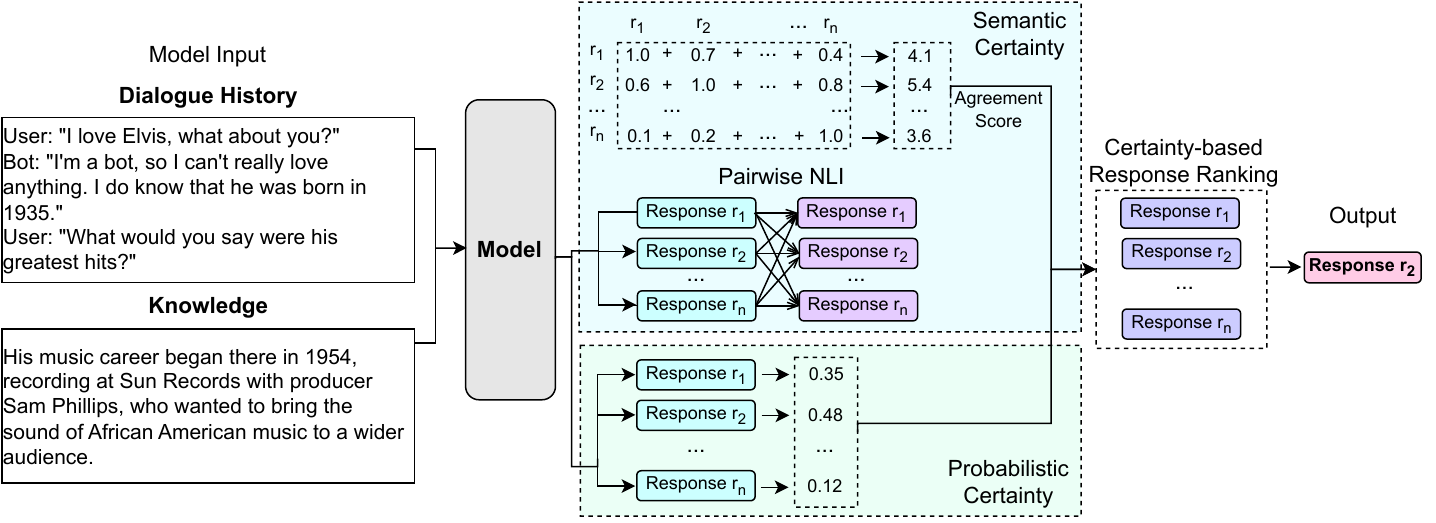}
\centering
\vspace{-1em}
\caption{\label{illustration} Illustration of the proposed Certainty-based Response Ranking approach. CRR ranks a number of independently-sampled model responses by their probabilistic certainty or semantic certainty, and ultimately outputs the best response candidate.}
% First, the pairwise consistency score between the final top k response candidates is calculated and summed to produce the agreement scores for each potential output. Then, the top k candidates are ranked based on their agreement scores and the candidate with highest agreement score is returned as the ultimate output response.}
\vspace{-0.6em}
\end{figure}

% \vspace{-1.2em}
\section{Sequence-Level Certainty}
\label{method}
\vspace{-0.8em}
This study proposes \textbf{sequence-level certainty} as a more general common theme across hallucination phenomena in KGDG task.
Different from previously proposed token-level certainty estimation approaches, sequence-level certainty measures certainty by considering an output sequence as a whole.
We further dissect sequence-level certainty into \textbf{probabilistic certainty} and \textbf{semantic certainty}.

% \subsection{Definition}
\vspace*{-0.5\baselineskip}
\subsection{Probabilistic Certainty}
\vspace*{-0.5\baselineskip}
We define probabilistic sequence-level certainty of a generated sequence to be the arithmetic mean log-probability of the entire sequence, as defined in previous works~\citep{kuhn2023semantic,murray2018correcting}.
Given a generated sequence \(\vs\) with length \(N\), the sequence-level probabilistic certainty can be calculated as: 
\(\frac{1}{N} \sum_{i=1}^N \text{log} \; p (s_i|\vs_{<i})\), where \(p (s_i | \vs_{<i})\) is the conditional probability of generating token \(s_i\) in sequence \(\vs\) given past tokens.
% This stands in contrast with the previously proposed probabilistic uncertainty measures, which only takes into account token-level information.

\vspace*{-0.5\baselineskip}
\subsection{Semantic Certainty}
\vspace*{-0.5\baselineskip}
We define semantic sequence-level certainty to be the level of confidence of a model generating the semantic contents of a response.
To estimate the certainty in meaning-space, we propose to use Agreement Score (AS) as a proxy of semantic certainty, which is explained below.
% For instance, suppose that we individually sample $n$ model outputs given the same context, among which $80\%$ consists of semantic contents that aligns with each other.
% Give the high confidence of the model to produce these responses with aligned semantic contents, we can say that these outputs have \textbf{high semantic certainty} level.
% In contrast, if semantic contents of the rest of the $20\%$ model outputs don't align with any other responses, then we can say that these outputs have \textbf{low semantic certainty} level.

\textbf{Agreement Score (AS)} \;
Given a context \(\vx\), we individually sample \(N\) model response candidates to constitute set \(\sS = \{\vs^{(1)}, \vs^{(2)}, ..., \vs^{(N)}\}\).
Let the relation \(\text{Entailment}(\cdot , \cdot)\) denote the probability that two generated sequences entail each other, or semantically support each other.
Then, the AS of model response \(\vs^{(i)}\) can be calculated as: \( AS(\vs^{(i)}) = \sum_{j=1}^{N} \text{Entailment}(\vs^{(i)}, \vs^{(j)})\)
% \begin{equation}
%     AS(\vs^{(i)}) = \sum_{j=1}^{n} \text{Entailment}(\vs^{(i)}, \vs^{(j)})
% \end{equation}
, which is the summed probability of semantic entailment between \(\vs^{(i)}\) and all other candidates.

% \vspace*{-0.5\baselineskip}
% \subsubsection{Sequence-level Certainty and Hallucination}
% We further conducted statistical testing to prove $2$ crucial relationships between sequence-level certainty and the level of hallucination:
% i) faithful model responses have \textbf{higher certainty} than hallucinated answers, and
% ii) A higher certainty is \textbf{positively and significantly correlated} with a lower level of hallucination.

\vspace{-1em}
\section{Certainty-based Response Ranking}
\vspace{-1em}
\subsection{Method}
\vspace{-0.6em}
Based on our categorization of sequence-level certainty, we further propose two types of Certainty-based Response Ranking (CRR) to mitigate model hallucination during decoding time: \textbf{Probabilistic CRR (P-CRR)} and \textbf{Semantic CRR (S-CRR)}.
% Both CRR method consist of $3$ steps.
% \begin{enumerate}
% \itemsep0em 
%     \item Given the same input $\vx$, we individually sample $n$ response candidates generated by a KGDG model : \(\vr^{(1)}, \vr^{(2)}, ..., \vr^{(n)}\).
%     \item Then, we calculate each response's certainty level.
%     For P-CRR, we measure the probablistic certainty score.
%     For S-CRR, we measure the semantic certainty score, estimated by AS.
%     \item Next, the generated responses are ranked based on their certainty level.
% \end{enumerate}
% Given the same input $\vx$, we individually sample $n$ response candidates generated by a KGDG model : \(\vr^{(1)}, \vr^{(2)}, ..., \vr^{(n)}\).
Given the same input, we individually sample several response candidates generated by a KGDG model.
Then, we calculate each response's probabilistic sequence-level certainty for P-CRR and semantic certainty for S-CRR.
% For P-CRR, we measure the probabilistic certainty score.
% For S-CRR, we measure the semantic certainty score, estimated by AS.
% Next, the generated responses are ranked based on their certainty level.
Eventually, the model ranks candidates based on their certainty level and outputs the response candidate with the highest certainty.
An illustration of CRR is demonstrated in Figure \ref{illustration}.

\vspace{-1em}
\subsection{Experiments}
% We design and conduct experiments to prove the effectiveness of our proposed CRR approaches in reducing model hallucination on KGDG task.
% We hereby provide details of our experiments.
\vspace{-0.6em}
\subsubsection{Experiment Setup}
\label{experiment-setup}
\vspace{-0.5em}

\textbf{Model Choices} \;
We experimented with 4 different KGDG models, which are fine-tuned from 4 different base models of different sizes and structures: GPT2-small, GPT2-medium~\citep{radford2019language}, T5-base~\citep{2020t5}, and OpenLlama~\citep{openlm2023openllama}.
Details for fine-tuning and inferencing KGDG models are in Appendix \ref{appendix:experiment-details}.
For calculating AS, we utilize an off-the-shelf RoBERTa-Large-based~\citep{liu2019roberta} Natural Language Inference (NLI) model~\citep{nie-etal-2020-adversarial}.
% For hallucination evaluation, we use an off-the-shelf hallucination classification model proposed by~\citep{dziri2022faithdial}.
For hallucination evaluation, we follow the method in~\citet{dziri2022faithdial} to use FaithCritic, an off-the-shelf RoBERTa-Large-based hallucination classification model.

\vspace{-0.5em}
\textbf{Baselines} \;
To prove the effectiveness of CRR for hallucination mitigation, we conduct experiments using $3$ decoding methods: Beam Search~\citep{bisiani1992beam}, Top-k Sampling~\citep{fan-etal-2018-hierarchical}, and Nucleus Sampling~\citep{Holtzman2019TheCC} with Top-k.
We also compare CRR with the uncertainty-aware beam search method proposed by \cite{xiao-wang-2021-hallucination}, which is most related to our approach.

\vspace{-0.5em}
\textbf{Datasets}\; We fine-tune the 4 KGDG models on FaithDial~\citep{dziri2022faithdial}'s training dataset.
% details of which are in Appendix \ref{appendix:experiment-details}.
Evaluations for baseline approaches and CRR methods are conducted on FaithDial, CMU-DoG~\citep{cmu_dog_emnlp18} and TopicalChat \citep{gopalakrishnan2019topical}'s test datasets.
% for exploring the generalizability of CRR to out-of-distribution data.
\vspace{-0.5em}

\textbf{Reported Metrics}
We first show statistical results to prove that higher probabilistic and semantic sequence-level certainties are significantly correlated with lower hallucination in model responses.
% We follow the hallucination evaluation method in~\citet{dziri2022faithdial} to use FaithCritic, an off-the-shelf RoBERTa-Large-based hallucination classification model.
For experiments on CRR, we report the \textbf{percentage of faithful responses} in experiments.

\vspace{-1em}
\subsection{Results}
\label{experiment-results}
\vspace{-0.6em}
\subsubsection{Sequence-level Certainty and Hallucination} 
We conducted statistical testing to prove that:
(1) faithful responses have \textbf{higher certainty} than hallucinated answers, and
(2) \textbf{higher certainty levels} positively and significantly correlate with \textbf{lower hallucination levels}.
Additional details for hypotheses testing are provided in Appendix \ref{appendix:certainty}.

% Table \ref{tab:hypothesis-1} shows that faithful outputs have higher certainty than hallucinated ones.
\textbf{Hypothesis 1} We conduct t-testing with the Alternative Hypothesis (\(H_1\)) being that faithful model responses have higher certainty levels than hallucinated ones, and Null Hypothesis (\(H_0\)) indicating no significant difference in certainty levels.
Results in Table \ref{tab:hypothesis-1} validates \(H_1\), indicating that \textbf{both probablistic and semantic sequence-level certainties are significantly higher in faithful outputs than in hallucinated ones.}.

\begin{table}[h]
\small
\begin{center}
%p{5.5cm} p{2.2cm} p{2cm} p{2cm} p{2cm}
\begin{tabular}{p{4.5cm}lllll} 
 \toprule 
 \multirow{2}{*}{\textbf{Hypothesis}} & \multirow{2}{*}{\textbf{Model}} & \multicolumn{2}{c}{\textbf{probabilistic}} & \multicolumn{2}{c}{\textbf{semantic}} \\
  \cmidrule{3-6}
 & & \textbf{p value} & \textbf{signif.} & \textbf{p value} & \textbf{signif.}\\
 \midrule
  \multirow{4}{*}{\parbox{2\linewidth}{
  Certainty (faithful responses)\\\tableblank{32pt}$>$\\ Certainty (hallucinated responses)}} & 
  GPT2-small & $\textbf{7.51E-210}$& $\checkmark$ & $\textbf{2.61E-148}$ & $\checkmark$  \\
 % $2.61E-148$ & Significant \\
  & GPT2-medium & $\textbf{2.32E-157}$& $\checkmark$ & $\textbf{7.51E-115}$ & $\checkmark$ \\
  % $7.51E-115$ & Significant\\
  & T5-base & $\textbf{9.17E-169}$ & $\checkmark$  & $\textbf{9.16E-19}$ & $\checkmark$  \\
  % 3.03E-59$ & Significant\\ 
  & OpenLlama-3B  & $\textbf{9.17E-169}$ & $\checkmark$ & $\textbf{2.79E-47}$ & $\checkmark$ \\
  % $2.79E-47$ & Significant\\
 \bottomrule
\end{tabular}
\end{center}
\vspace{-1em}
\caption{\label{tab:hypothesis-1} Experiment results. 
% Significant p values are in bold. 
Across all $4$ models, levels of both probabilistic certainty and semantic certainty of faithful model responses are significantly higher than that of hallucinated responses.}
\vspace{-1em}
\end{table}

\textbf{Hypothesis 2} We use the Point-Biserial Correlation Coefficient (PBCC) to show correlation between certainty level and probability of hallucination for response candidates.
Table \ref{tab:hypothesis-2} shows that both types of certainty in model responses are negatively and significantly correlated with the probability of hallucination, meaning that \textbf{higher sequence-level certainty corresponds to lower hallucination}.
\vspace{-0.6em}

\begin{table}[h]
\small
\begin{center}
\begin{tabular}{p{2.2cm} p{1.5cm} p{4.2cm} p{4.2cm} } 
 \toprule 
 \multirow{2}{*}{\textbf{Model}} &  \multirow{2}{*}{\textbf{\# Params}} & \multicolumn{2}{c}{\textbf{Point-Biserial Correlation Coeff. with Hallucination Probability}} \\
\cmidrule{3-4}
& & \textbf{Probabilistic Certainty} & \textbf{Semantic Certainty}\\ 
\midrule
\textbf{GPT2-small} & $117$M & $-0.265$ (p-value $\ll \textbf{0.01}$) & $-0.165$ (p-value $\ll \textbf{0.01}$) \\
% $-0.16$  &  $8.72E-96$ & $\checkmark$  & $\ll 0.01$ \\
\textbf{GPT2-medium} & $345$M & $-0.231$ (p-value $\ll \textbf{0.01}$) & $-0.146$ (p-value $\ll \textbf{0.01}$) \\
%  & $-0.14$ & $6.89E-77$ & $\checkmark$ & $\ll 0.01$ \\ 
\textbf{T5-base} & $220$M & $-0.205$ (p-value $\ll \textbf{0.01}$) & $-0.067$ (p-value $\ll \textbf{0.01}$) \\
% $-0.12$ & $1.24E-58$ & $\checkmark$ & $\ll 0.01$ \\
\textbf{OpenLlama-3B} & $3$B & $-0.173$ (p-value $\ll \textbf{0.01}$) & $-0.110$ (p-value $\ll \textbf{0.01}$) \\
% $-0.10$ & $1.34E-38$ & $\checkmark$ & $\ll 0.01$ \\
 \bottomrule
\end{tabular}
\end{center}
\vspace{-1em}
\caption{\label{tab:hypothesis-2} Experiment results. Both types of sequence-level certainty are negatively and significantly correlated with hallucination probability, as measured by Point-Biserial Correlation..}
\vspace{-0.6em}
\end{table}

\vspace{-1em}
\subsubsection{Effectiveness of CRR for Hallucination Mitigation}
Table \ref{tab:results-1} shows experiment results using different hallucination mitigation methods on GPT2-small.
%  using different decoding methods.
% We report the percentage of faithful model responses across $3$ datasets and $3$ decoding methods.
% Across all datasets and decoding methods, w
Both P-CRR and S-CRR improve response faithfulness.
% In addition, P-CRR achieves better performance than S-CRR in most cases.
Among different decoding methods, Nucleus Sampling with Top-k and P-CRR achieves the best performance, with $97.6 \%$ faithful generations.
What's more, \citeauthor{xiao-wang-2021-hallucination}'s method (row 2) \footnote{Note that since \citeauthor{xiao-wang-2021-hallucination}'s method is specifically designed for beam search, it cannot be applied to other decoding methods.} fails to achieve faithfulness improvement, indicating that controlling token-level uncertainty cannot effectively reduce hallucination on KGDG.
% Through comparison between CRR and other hallucination mitigation approaches, we see that ....
% \vspace{-1em}

\begin{table}[h]
\small
\begin{center}
\begin{tabular}{llccc} 
 \toprule 
\multirow{2}{*}{\textbf{Decoding Method}} & \multirow{2}{*}{\textbf{Mitigation Method}}  & \multicolumn{3}{c}{\textbf{Dataset}} \\
 \cmidrule{3-5}
 & & \textbf{FaithDial $\uparrow$} & \textbf{CMU-DoG $\uparrow$} & \textbf{TopicalChat $\uparrow$} \\
 \midrule
 \textbf{Beam Search} & None & $66.0$ & $43.2$ & $12.1$\\
 & Uncertainty-Aware & $65.0$ & $43.7$ & $13.2$\\
  &  P-CRR & $\textbf{73.9}$ & $42.9$ & $11.6$ \\
 & \textbf{S-CRR} & $71.6$ & $\textbf{44.8}$ & $\textbf{13.2}$ \\ 
\midrule
 \textbf{Top-k Sampling} & None & $83.4$ & $32.1$ & $12.4$\\
 & \textbf{P-CRR} & $\textbf{95.6}$ & $\textbf{46.3}$ & $\textbf{16.5}$ \\
 & S-CRR & $89.9$ & $34.2$ & $14.3$\\
\midrule
 \textbf{Nucleus Sampling} & None & $91.2$ & $38.3$ & $14.6$\\  
 & \textbf{P-CRR} & $\textbf{97.6}$ & $\textbf{50.0}$ & $\textbf{16.7}$ \\
 & S-CRR & $95.7$ & $40.8$ & $15.1$\\
 \bottomrule 
\end{tabular}
\end{center}
% \caption{Experiment results on GPT2-small with different hallucination mitigation approaches across $3$ datasets, with number of return sequences set to $5$. Percentage of faithful model responses are reported for all methods. Best-performing method and reported score are bolded.}
\vspace{-1em}
\caption{\label{tab:results-1} Experiment results on GPT2-small with different decoding methods across $3$ datasets. Faithful percentages of responses are reported. Best-performing methods and scores are bolded.}
\vspace{-1em}
\end{table}
% \vspace*{-0.3\baselineskip}

\vspace{-0.6em}
\textbf{Generalizability To Different Models} \;
Table \ref{tab:results-2} shows experiment results on GPT2-medium, T5-base, and OpenLlama-3B to show the generalizability of CRR to different KGDG models.
% models across $3$ datasets.
% Through comparison between CRR and the baselines across all models and all $3$ decoding methods, we observe that ...
% Results on these $3$ models demonstrate similar 
Similar to trends in Table \ref{tab:results-1}, both P-CRR and S-CRR achieve significant improvements in faithful response percentages over the baselines in most settings.
% , while P-CRR always achieves better performance.

\begin{table}[htb]
\vspace{-0.3em}
\small
\begin{center}
%p{2.5cm}p{2.6cm}p{2cm}p{2cm}p{2.1cm}
\begin{tabular}{llcccc} 
 \toprule 
 \multirow{2}{*}{\textbf{Base Model}} & \multirow{2}{*}{\textbf{\# Params}} & \multirow{2}{*}{\textbf{Decoding}} & \multicolumn{3}{c}{\textbf{Dataset}} \\
 \cmidrule{4-6}
 & & & \textbf{FaithDial $\uparrow$} & \textbf{CMU-DoG $\uparrow$} & \textbf{TopicalChat $\uparrow$} \\
 \midrule
\multirow{9}{*}{\textbf{GPT2-medium}} & \multirow{9}{*}{$345$M} & Beam Search & $71.6$ & $43.3$ & $14.8$\\
& & \;\; \;\; + P-CRR & $77.1$ & $45.0$ & $14.9$\\
& &\textbf{\;\;\;\;  + S-CRR} & $\textbf{77.4}$ & $\textbf{47.1}$ & $\textbf{16.0}$ \\
\cmidrule{3-6}
 & & Top-k Sampling  & $87.3$ & $36.5$ & $15.3$\\
 &  & \;\; \;\;  \textbf{+ P-CRR} & $\textbf{96.9}$ & $\textbf{49.5}$ & $\textbf{18.9}$\\
 & & \;\; \;\;  + S-CRR & $92.6$ & $41.7$ & $17.1$\\
   \cmidrule{3-6}
 & & Nucleus Sampling & $93.8$ & $43.4$ & $17.0$\\
& & \;\; \;\;  \textbf{+ P-CRR} & $\textbf{98.2}$ & $\textbf{53.4}$ & $\textbf{19.5}$\\
 & & \;\; \;\;  + S-CRR & $96.8$ & $48.6$ & $18.2$\\
  \midrule
\multirow{9}{*}{\textbf{T5-Base}} & \multirow{9}{*}{$220$M} & Beam Search & $99.3$ & $66.1$ & $\textbf{27.0}$\\
& & \;\; \;\;  \textbf{+ P-CRR} & $\textbf{99.5}$ & $\textbf{67.1}$ & $25.4$\\
 & & \;\; \;\;  + S-CRR & $99.4$ & $\textbf{67.1}$ & $26.6$ \\
\cmidrule{3-6}
&  & Top-k Sampling  & $78.8$ & $38.7$ & $23.2$\\
& & \;\; \;\;  \textbf{+ P-CRR} & $\textbf{91.6}$ & $\textbf{48.1}$ & $\textbf{25.2}$\\
 & & \;\; \;\;  + S-CRR & $80.2$ & $42.0$ & $23.7$\\
   \cmidrule{3-6}
&  & Nucleus Sampling & $87.7$ & $47.6$ & $26.4$\\
& & \;\; \;\;  \textbf{+ P-CRR} & $\textbf{95.3}$ & $\textbf{54.8}$ & $23.8$ \\
& & \;\; \;\;  + S-CRR & $89.1$ & $52.4$ & $\textbf{27.9}$\\
 \midrule
\multirow{9}{*}{\textbf{OpenLlama-3B}} & \multirow{9}{*}{$3$B} & Beam Search & $68.3$ & $\textbf{44.1}$ & $\textbf{17.4}$\\
 & & \;\; \;\;  + P-CRR & $70.0$ & $41.9$ & $16.4$ \\
 & & \textbf{\;\; \;\;  + S-CRR} & $\textbf{75.3}$ & $40.6$ & $16.1$ \\
\cmidrule{3-6}
&  & Top-k Sampling  & $90.9$ & $39.1$ & $21.9$\\  
&  & \;\; \;\;  \textbf{+ P-CRR} & $\textbf{97.4}$ & $\textbf{50.4}$ & $\textbf{25.1}$\\
&  & \;\; \;\;  + S-CRR & $94.4$ & $42.5$ & $22.9$\\
 \cmidrule{3-6}
&  & Nucleus Sampling & $95.7$ & $45.1$ & $23.5$\\
&  & \;\; \;\; \textbf{+ P-CRR} & $\textbf{98.6}$ & $\textbf{53.5}$ & $\textbf{25.9}$ \\
& & \;\; \;\;  + S-CRR & $97.2$ & $48.1$ & $23.7$\\
 \bottomrule 
\end{tabular}
\vspace{-0.8em}
\end{center}
\caption{\label{tab:results-2}Experiment results for the baselines and the proposed CRR approaches. Faithfulness percentage is reported for all methods. Best-performing method and reported score are in bold.}
% \vspace{-1.2em}
\end{table}

\textbf{Generalizability To Different Number of Response Candidates} \;
We conduct ablation experiments to investigate the generalizability of CRR when different numbers of response candidates are sampled.
Table \ref{tab:num-return-sequences} shows results on GPT2-small on FaithDial's test set when $5$, $10$, and $20$ response candidates are sampled during response ranking.
% ith $3$ different numbers of response candidates during response ranking: $5$, $10$, and $20$.
S-CRR achieves significant performance improvement with an increase in the number of response candidates.
%  sampled for ranking.
This indicates that by aligning more candidates with each other, S-CRR better captures the semantic certainty of each response.
%  in the semantic contents of each candidate.
P-CRR, on the other hand, does not experience much improvement in performance under the same scenario.
% This observation further shows that S-CRR can capture semantic certainty in model responses.

\begin{table}[htb]
\small
\begin{center}
%p{2.5cm}p{2.6cm}p{2cm}p{2cm}p{2.1cm}
\begin{tabular}{llccc} 
 \toprule 
\multirow{2}{*}{\textbf{Decoding Method}} & \multirow{2}{*}{\textbf{Mitigation Method}} & \multicolumn{3}{c}{\textbf{FaithDial} $\uparrow$} \\
 \cmidrule{3-5}
 & & \textbf{$\#$ seq 5} & \textbf{$\#$ seq 10} & \textbf{$\#$ seq 20} \\
 \midrule
\textbf{Beam Search} & None & $66.0$ & $66.0$ & $66.0$ \\
& Uncertainty-Aware & $65.0$ & $65.0$ & $65.0$ \\
 & \textbf{P-CRR} & $\textbf{73.9}$ & $73.7$ & $72.9$ \\
 & \textbf{S-CRR} & $71.6$ & $\textbf{76.6}$ & $\textbf{78.1}$\\ 
 \midrule
 \textbf{Top-k Sampling} & None & $83.4$ & $83.4$ & $83.4$ \\
  & \textbf{P-CRR} & $\textbf{95.6}$ & $\textbf{96.5}$ & $\textbf{97.3}$ \\
  & S-CRR & $89.9$ & $93.2$ & $94.5$ \\
 \midrule
 \textbf{Nucleus Sampling} & None & $91.2$ & $91.2$ & $91.2$ \\
 & \textbf{P-CRR} & $\textbf{97.6}$ & $\textbf{98.4}$ & $\textbf{98.7}$ \\
 & S-CRR & $95.7$ & $97.1$ & $97.7$\\
 \bottomrule 
\end{tabular}
\end{center}
\caption{Experiment results.
% for the generalizability of CRR to different numbers of response candidates using GPT2-small on FaithDial. 
Note that since the original decoding methods and uncertainty-aware beam search do not rank sampled responses, their reported scores are invariant to the number of sampled response candidates.}
\label{tab:num-return-sequences}
\end{table}
\vspace*{-0.5\baselineskip}

% \vspace{-1.2em}
\section{Background on Uncertainty and Hallucination} 
% Previous research has explored and proposed different methods to measure uncertainty in model predictions~\citep{xiao-wang-2021-hallucination, kuhn2023semantic, manakul2023selfcheckgpt}.
% Among these works,~\citeauthor{xiao-wang-2021-hallucination} was the first to explore the relationship between token-level predictive uncertainty and hallucination on the Image Captioning (IC) task.
% However, their proposed uncertainty estimation did not consider sequence-level or semantic-level information in generated text.
% ~\citeauthor{kuhn2023semantic} further extended the estimation of model uncertainty to semantic space on the QA task, but did not explore how semantic uncertainty measurement is related to level of hallucination.
% Our study draws inspirations from these works, and is among the first to comprehensively and systematically explore the relationship between both sequence-level probabilistic and semantic certainty and level of hallucination in generative models.
% In this section, we briefly introduce the methods proposed by previous works to estimate uncertainty, as well as provide the background on the correlation between uncertainty and hallucination, as observed by prior studies.

\vspace{-0.6em}
\subsection{Uncertainty Estimation in Generative Models}
\vspace{-0.6em}
Previous researchers~\citep{xiao-wang-2021-hallucination, kuhn2023semantic, zhang2023rtuning, liu2024examining} have studied probabilistic uncertainty and semantic uncertainty, but mainly in different contexts from hallucination.
\citeauthor{xiao-wang-2021-hallucination} proposed token-level predictive uncertainty, which 
% quantifies the entropy of the token probability distribution during generation.
% that a model predicts in language generation.
% Their work adopts a probability-based approach and 
formulates the total predictive uncertainty of a predicted token as its entropy.
% We denote this method as the \textbf{probabilistic uncertainty} estimation.
~\citeauthor{kuhn2023semantic}'s work extends the exploration of uncertainty to the semantic aspect.
They establish \textbf{semantic uncertainty} as the entropy of the random variable representing the output distribution in the semantic event-space, and explored how it is predictive of model accuracy on Question Answering (QA) tasks.
% , and did not explore its relationship with hallucination.

\vspace{-0.9em}
\subsection{On Uncertainty and Hallucination}
\vspace{-0.6em}
\citeauthor{xiao-wang-2021-hallucination}'s work was the first to explore the correlation between model uncertainty and hallucination.
They observed that on the Image Captioning (IC) task, higher token-level probabilistic uncertainty corresponds to a higher chance of hallucination.
% Based on this observation, they further proposed uncertainty-aware beam search, which takes into account the predictive uncertainty during beam search to reduce hallucination in model-predicted captions for images.
They also proposed uncertainty-aware beam search, which accounts for the token-level uncertainty during generation to reduce hallucination.
However, their experiments were limited to token-level uncertainty in IC tasks.
Additionally, their uncertainty-aware beam search cannot be applied to other decoding methods such as top-k sampling.
\citeauthor{manakul2023selfcheckgpt}'s work showed that probability-based model uncertainty can be used to detect hallucinations on QA tasks.
However, they neither provide insights on the relationship between uncertainty and hallucination, nor propose mitigation solutions.

\vspace*{-1\baselineskip}
\section{Conclusion}
\vspace*{-0.6\baselineskip}
 In this paper, we explore the relationship between \textbf{sequence-level certainty} and hallucination in KGDG.
 We dissect sequence-level certainty in model generation into \textbf{probabilistic certainty} and \textbf{semantic certainty}. 
 Probabilistic certainty measures the statistical likelihood of generating a sequence, whereas semantic certainty measures the probability of generating specific semantic contents in a response.
 % Through experimenting on KGDG task with $4$ models of various structures and sizes, we show that both a higher level of probabilistic certainty and a higher level of semantic certainty are positively and significantly correlated with a lower level of hallucination in model response.
 % Furthermore, based on our empirical observations, w
 Furthermore, we propose Certainty-based Response Ranking (CRR), a decoding-time method to mitigate hallucination in model generations by outputting candidate responses with the highest certainty levels.
 Based on our categorization of certainty, we propose Probabilistic CRR (P-CRR) and Semantic CRR (S-CRR) to address hallucinations from different perspectives.
 Through experimenting on $4$ models across $3$ decoding methods on $3$ datasets, we prove the effectiveness of both P-CRR and S-CRR in reducing model hallucination on the KGDG task.

\appendix
% \section{Relationship Between Sequence-Level Certainty and Hallucination}
\section{Experimental Details}
\label{appendix:experiment-details}
\subsection{Task Definition}
For the KGDG task, a model is provided with a series of dialogue history and a piece of textual knowledge, and is required to generate a response to the dialogue history according to the given knowledge.
Responses generated by a faithful KGDG model should be truthful to the knowledge provided in its input.

\subsection{Training and Inferencing KGDG Models}
We conduct experiments on KGDG models to validate the relationship between sequence-level certainty and hallucination, and to test the proposed CRR method for hallucination mitigation.
Following the method used in previous work~\citep{dziri2022faithdial}, we select $4$ base models to and further fine-tuned them on the KGDG task to build KGDG models.
% Then, we explore the correlation between the $2$ types of sequence-level certainty and the level of hallucination in model response candidates.
Training details are provided below.

\textbf{Model Selection} \;
As mentioned in Section \ref{experiment-setup}, we select 4 different base models of different sizes and structures as base models: GPT2-small, GPT2-medium~\citep{radford2019language}, T5-base~\citep{2020t5}, and OpenLlama~\citep{openlm2023openllama}.
% For calculating the pairwise entailment probability during agreement score calculation, we utilize an off-the-shelf RoBERTa-Large-based~\citep{liu2019roberta} Natural Language Inference (NLI) model~\citep{nie-etal-2020-adversarial} and utilize the probability of entailment between pairs of response candidates.
% For hallucination evaluation, we use an off-the-shelf RoBERTa-Large-based hallucination classification model trained on FaithCritic, which is a derivative of the FaithDial dataset~\citep{dziri2022faithdial}.
% Given a piece of knowledge and a model response, the hallucination classification model classifies the model's output as either being ``hallucinated'' or ``faithful''.

\textbf{Dataset} \;
For training the KGDG models, we utilize FaithDial~\citep{dziri2022faithdial}, a faithful knowledge-grounded dialogue corpus built from the Wizard of Wikipedia dataset~\citep{dinan2019wizard}.
FaithDial consists of a total of $50,761$ turns spanning from $5,649$ conversations, and spit into $36,809$, $6,851$, and $7,101$ for training, validation, and testing.
% We utilize the full training, validation, and test sets of FaithDial for training, selecting, and evaluating the KGDG model. 

\textbf{Training Details} \;
Following hyper-parameter settings in \cite{dziri2022faithdial}, we train the KGDG models for $10$ epochs with batch size set to $16$ and maximum sequence length set to $512$.
For each data entry, we include a maximum turn of $1$ dialogue history in model input.
For optimization, we use linear scheduler for the AdamW optimizer~\citep{Loshchilov2017DecoupledWD}, with learning rate set to $6.25\times10^{-5}$, warmup ratio set to $0.04$, epsilon set to $1\times10^{-8}$, and weight decay set to $0$.
Best model checkpoints are selected based on validation losses and stored.
% \textbf{Inference Details} \;
% During inference time, we select nucleus + top-k sampling decoding method for generation.
% We set temperature to $1.0$, top k to $50$, top p to $0.9$, and maximum new tokens to $100$.
% All hyper-parameters for generation are selected to ensure best possible quality of generated text.

\section{Inference-Time Decoding Methods}
% \subsubsection{Implementation}
% \textbf{Model Details} \; Following experimental settings in Section \ref{method}, we conduct evaluation on 4 different base models of different sizes and structures: GPT2-small, GPT2-medium~\citep{radford2019language}, T5-base~\citep{2020t5}, and OpenLlama~\citep{openlm2023openllama}.
% All models are fine-tuned for KGDG task on the FaithDial~\citep{dziri2022faithdial} dataset, as described in Section \ref{method}.

% \textbf{Datasets} \; In addition to evaluation on FaithDial~\citep{dziri2022faithdial}'s test dataset, we also want to explore the generalizability of CRR to out-of-distribution data.
% Therefore, we follow previous works~\citep{dziri2022faithdial} also utilize the test sets of $2$ additional KGDG datasets in our experiments: CMU-DoG~\citep{cmu_dog_emnlp18} and TopicalChat \cite{gopalakrishnan2019topical}.
Below, we provide details for implementing different decoding methods at inference time.
For all decoding methods, we set the maximum number of new tokens to $100$.

\textbf{Baseline Decoding Methods} \;
As discussed in Section \ref{experiment-setup}, we experiment with $3$ different baseline decoding methods at inference time: Beam Search~\citep{bisiani1992beam}, Top-k Sampling~\citep{fan-etal-2018-hierarchical}, and Nucleus Sampling~\citep{Holtzman2019TheCC} with Top-k.
The beam search method in experiments is based on our implementation of the decoding algorithm.
For beam search decoding, we set the beam size to $5$.
For top-k sampling decoding, we set the temperature to $1.0$, and top k to $50$.
For nucleus sampling with top-k, we set the temperature to $1.0$, top-k to $50$, and top-p to $0.9$.

\textbf{Ablation Study Methods} \;
We also implement the Uncertainty-Aware Beam Search method proposed by \citet{xiao-wang-2021-hallucination} to establish a comparison with the proposed CRR methods.
Since \citet{xiao-wang-2021-hallucination}'s proposed approach was originally designed for image captioning tasks, experiments in our paper are based on our modified implementation of the method on KGDG task.
Following the setting in \citet{xiao-wang-2021-hallucination}'s implementation, we set the uncertainty lambda to $0.2$ when considering the epistemic uncertainty of the model during beam search.

\textbf{CRR Methods} \; 
For both CRR methods, we choose to sample and rank $5$ response candidates for each input.
As mentioned in Section \ref{experiment-setup}, an off-the-shelf NLI model~\citep{nie-etal-2020-adversarial} is used to calculate the AS between output candidates at inference time.

\section{Relationship Between Sequence-Level Certainty and Hallucination}
\label{appendix:certainty}
% We design and conduct experiments on the KGDG task to further explore the correlation between sequence-level certainty and level of hallucination.
In Section \ref{experiment-results}, we demonstrated that (1) faithful model responses have significantly higher certainty than hallucinated answers, ane (2) a higher certainty is positively and significantly correlated with a lower level of hallucination using the Point-Biserial Correlation Coefficient,
% Below, we first provide additional details for the correlation experiment, and then introduce an additional experiment to show 
Below, we provide additional details for the statistical testing experiments.

\subsection{Details for Proving Hypothesis 1}
\label{appendix:pb-details}
% Certainty of Faithful vs. Hallucinated Responses} 
We show that faithful model responses demonstrate higher levels of sequence-level certainty than unfaithful answers.
% For each input in FaithDial's test dataset,
% we individually sample $5$ candidate responses from a KGDG model.
% Then, we calculate the sequence-level probabilistic and semantic certainty, as well as the probability of hallucination for each of the response candidates.
For evaluation data, we first generate responses on FaithDial~\citep{dziri2022faithdial}'s test set.
For each data entry, we individually sample $5$ candidate responses.
We select the nucleus + top-k sampling decoding method for generation, setting the temperature to $1.0$, top k to $50$, top p to $0.9$, and maximum new tokens to $100$.
All hyper-parameters for generation are selected to ensure the best possible quality of the generated text.
% we use the same set of generated responses on FaithDial's test set as in Section \ref{appendix:pb-details}.
We classify the faithfulness of each response using FaithCritic~\citep{dziri2022faithdial}, and calculate their probabilistic and semantic sequence-level certainties.
We then conduct t-testing with the Null Hypothesis (\(H_0\)) being that faithful model responses don't have higher certainties than hallucinated responses, and the Alternative Hypothesis (\(H_1\)) being that faithful model responses have higher certainty levels than hallucinated ones.
P-values and levels of significance are reported.

\subsection{Details for Proving Hypothesis 2}
%  Point-Biserial Correlation Experiment} 
% \label{appendix:pb-details}
We use the same set of generated responses on FaithDial's test set as in Section \ref{appendix:pb-details} to investigate the correlation between sequence-level certainty of model responses and the level of hallucination.
% we first generate responses on FaithDial~\citep{dziri2022faithdial}'s test set.
% For each data entry, we individually sample $5$ candidate responses.
% We select the nucleus + top-k sampling decoding method for generation, setting the temperature to $1.0$, top k to $50$, top p to $0.9$, and maximum new tokens to $100$.
% All hyper-parameters for generation are selected to ensure the best possible quality of the generated text.
For each response, we calculate the probabilistic and semantic sequence-level certainties, and obtain the probability of hallucination of each response using FaithCritic~\citet{dziri2022faithdial}.
% We choose to use PBCC in our experiment 
Since we establish hallucination detection as a binary classification task and certainty level as continuous values, we choose to report the Point-Biserial Correlation Coefficient (PBCC) between the two types of sequence-level certainty and the probability of hallucination in response candidates.
We also show the level of significance for the PBCC tests.

\section{Case Study: Effectiveness of CRR}
\subsection{Case Study Using S-CRR}
Table \ref{tab:visualization-scrr} demonstrates $2$ case studies of model responses using the original nucleus sampling with top-k decoding vs. using Semantic CRR for response ranking.
 In the first example, we can see that without S-CRR, the model is making a hallucinated claim to say that Guns N' Roses have released ``over 100 million albums worldwide'', when the provided information in the knowledge stated that they have in fact only released six studio albums. 
 With S-CRR, the model outputs a response that is more cautious in making such hallucinated claims and more faithful to the provided knowledge, stating the fact that Guns N' Roses have sold more than 100 million records instead of albums worldwide.

\begin{table}[hbt]
\begin{center}
\begin{tabular}{p{3cm} p{10cm}} 
\hline \hline
 \multicolumn{2}{c}{\textbf{Input 1}} \\ [0.5ex] 
 \hline \hline 
 % \textbf{Dialogue History} & User: I love candy, what's a good brand?  \\ [1.5ex] 
  \textbf{Knowledge} & Guns N' Roses has released six studio albums, accumulating sales of more than 100 million records worldwide, including 45 million in the United States, making them the 41st best-selling artist of all time. \\ [1.5ex]
%  \textbf{Dialogue History} & \textbf{Knowledge} &  \textbf{Top-k Response} & \textbf{ARR Response} \\ [0.5ex] 
 \hline \hline
 \multicolumn{2}{c}{\textbf{Model Response}} \\ [0.5ex] 
 \hline \hline
\textbf{Nucleus Sampling} &  I know that they have released  \textcolor{red}{over 100 million albums worldwide.} \\ [0.5ex] 
\hline
 % \textbf{ \:\:\:\:\:\:\:\:\:\:\:\:\:\:\:\:\:\:\: +P-CRR} &   \\[0.5ex] 
 % \hline
 \textbf{ \:\:\:\:\:\:\:\:\:\:\:\:\:\:\:\:\:\:\: +S-CRR} &  I have, and have heard of them, which have sold \textcolor{forest}{more than 100 million records worldwide.} \\[0.5ex] 
 \hline \\ [0.5ex]  
 \hline \hline
 \multicolumn{2}{c}{\textbf{Input 2}} \\ [0.5ex] 
 \hline \hline 
 % \textbf{Dialogue History} & User: I love candy, what's a good brand?  \\ [1.5ex] 
  \textbf{Knowledge} & Presley was born in Tupelo, Mississippi, and relocated to Memphis, Tennessee, with his family when he was 13 years old. \\ [1.5ex]
%  \textbf{Dialogue History} & \textbf{Knowledge} &  \textbf{Top-k Response} & \textbf{ARR Response} \\ [0.5ex] 
 \hline \hline
 \multicolumn{2}{c}{\textbf{Model Response}} \\ [0.5ex] 
 \hline \hline
\textbf{Nucleus Sampling} &  He was born in \textcolor{red}{Memphis, Tennessee.} \\ [0.5ex] 
\hline
 % \textbf{ \:\:\:\:\:\:\:\:\:\:\:\:\:\:\:\:\:\:\: +P-CRR} &   \\[0.5ex] 
 % \hline
 \textbf{ \:\:\:\:\:\:\:\:\:\:\:\:\:\:\:\:\:\:\: +S-CRR} &  He was born in \textcolor{forest}{Mississippi.} \\[0.5ex] 
 \hline
\end{tabular}
\end{center}
\caption{Example of model responses using the original Nucleus + Top-k Sampling decoding method and response using the same decoding method with S-CRR. Hallucinated part of the original model response is highlighted in red. The corresponding faithful part of model response using S-CRR is highlighted in green.}
\label{tab:visualization-scrr}
\end{table}

\subsection{Case Study Using P-CRR}
Table \ref{tab:visualization-pcrr} demonstrates $2$ case studies of model responses using the original nucleus sampling with top-k decoding vs. using Probabilistic CRR for response ranking.
 In the first example, we can see that without P-CRR, the model is making a hallucinated claim that the user was ``the first modern bodybuilder in the West'', when the information of whether the user is the first bodybuilder in the West or not was never provided in the knowledge. 
 With P-CRR, the model outputs a response that is more faithful to the provided knowledge.

\begin{table}[htb]
\begin{center}
\begin{tabular}{p{3cm} p{10cm}} 
\hline \hline
 \multicolumn{2}{c}{\textbf{Input 1}} \\ [0.5ex] 
 \hline \hline 
 % \textbf{Dialogue History} & User: I love candy, what's a good brand?  \\ [1.5ex] 
  \textbf{Knowledge} & In the 1910s, Eugen Sandow, widely considered to be the first modern bodybuilder in the West, advocated the use of dietary control to enhance muscle growth. \\ [1.5ex]
%  \textbf{Dialogue History} & \textbf{Knowledge} &  \textbf{Top-k Response} & \textbf{ARR Response} \\ [0.5ex] 
 \hline \hline
 \multicolumn{2}{c}{\textbf{Model Response}} \\ [0.5ex] 
 \hline \hline
\textbf{Nucleus Sampling} &  Interesting, so \textcolor{red}{you were actually the first modern bodybuilder in the West}, did you have dietary control?. \\ [0.5ex] 
\hline
 % \textbf{ \:\:\:\:\:\:\:\:\:\:\:\:\:\:\:\:\:\:\: +P-CRR} &   \\[0.5ex] 
 % \hline
 \textbf{ \:\:\:\:\:\:\:\:\:\:\:\:\:\:\:\:\:\:\: +P-CRR} &  Well, did you know that in the 1910s, \textcolor{forest}{Eugen Sandow was considered the first modern bodybuilder?}\\[0.5ex] 
 \hline \\ [0.5ex]  
 \hline \hline
 \multicolumn{2}{c}{\textbf{Input 2}} \\ [0.5ex] 
 \hline \hline 
 % \textbf{Dialogue History} & User: I love candy, what's a good brand?  \\ [1.5ex] 
  \textbf{Knowledge} & Presley's first RCA single, ''Heartbreak Hotel'', was released in January 1956 and became a number one hit in the United States. \\ [1.5ex]
%  \textbf{Dialogue History} & \textbf{Knowledge} &  \textbf{Top-k Response} & \textbf{ARR Response} \\ [0.5ex] 
 \hline \hline
 \multicolumn{2}{c}{\textbf{Model Response}} \\ [0.5ex] 
 \hline \hline
\textbf{Nucleus Sampling} &  I don't have many, but \textcolor{red}{his first single came out in 1956} and hit number one in the US. \\ [0.5ex] 
\hline
 % \textbf{ \:\:\:\:\:\:\:\:\:\:\:\:\:\:\:\:\:\:\: +P-CRR} &   \\[0.5ex] 
 % \hline
 \textbf{ \:\:\:\:\:\:\:\:\:\:\:\:\:\:\:\:\:\:\: +P-CRR} & I don't know about his usual fans, but I do know that \textcolor{forest}{his first RCA single, ``Heartbreak Hotel'', was released in 1956.} \\[0.5ex] 
 \hline
\end{tabular}
\end{center}
\caption{Example of model responses using the original Nucleus + Top-k Sampling decoding method and response using the same decoding method with P-CRR. Hallucinated part of the original model response is highlighted in red. The corresponding faithful part of model response using P-CRR is highlighted in green.}
\label{tab:visualization-pcrr}
\end{table}

\end{document}

%% file: math_commands.tex
\usepackage{amsmath,amsfonts,bm}

\def\eqref#1{equation~\ref{#1}}
\def\1{\bm{1}}

\def\vs{{\bm{s}}}

\def\vx{{\bm{x}}}

\DeclareMathAlphabet{\mathsfit}{\encodingdefault}{\sfdefault}{m}{sl}
\SetMathAlphabet{\mathsfit}{bold}{\encodingdefault}{\sfdefault}{bx}{n}

\def\sS{{\mathbb{S}}}